\documentclass[11pt]{article}
\usepackage{hyperref}
\usepackage{acl2016}
\usepackage{natbib}
\usepackage{times}
\usepackage{latexsym}
\usepackage{amsmath}
\usepackage{amssymb}
\usepackage{graphicx}
\usepackage{ifthen}
\usepackage[utf8]{inputenc}
\usepackage{tabularx}
\usepackage{multirow}
\usepackage{multicol}
\usepackage{booktabs}
\usepackage{arydshln}
\usepackage{enumitem}
\usepackage{color}
\usepackage{listings}
\usepackage{csquotes}
\usepackage{url}
\usepackage{subcaption}
\usepackage[noline,boxed]{algorithm2e}
\usepackage[T1]{fontenc}
\usepackage[polish,english]{babel}
\usepackage[skip=0.2em]{caption}

\newcommand\sY{\ensuremath{\mathcal{Y}}}
\newcommand\sZ{\ensuremath{\mathcal{Z}}}

\newcommand\R{\ensuremath{\mathbb{R}}} 
\newcommand\eqdef{\ensuremath{\stackrel{\rm def}{=}}} 
\newcommand{\1}{\mathbb{I}} 
\newcommand\refeqn[1]{(\ref{eqn:#1})}

\newcommand\refsec[1]{Section~\ref{sec:#1}}

\newcommand\reffig[1]{Figure~\ref{fig:#1}}

\newcommand\reftab[1]{Table~\ref{tab:#1}}

\newcommand\refalg[1]{Algorithm~\ref{alg:#1}}

\ifthenelse{\isundefined{\definition}}{\newtheorem{definition}{Definition}}{}
\ifthenelse{\isundefined{\assumption}}{\newtheorem{assumption}{Assumption}}{}
\ifthenelse{\isundefined{\hypothesis}}{\newtheorem{hypothesis}{Hypothesis}}{}
\ifthenelse{\isundefined{\proposition}}{\newtheorem{proposition}{Proposition}}{}
\ifthenelse{\isundefined{\theorem}}{\newtheorem{theorem}{Theorem}}{}
\ifthenelse{\isundefined{\lemma}}{\newtheorem{lemma}{Lemma}}{}
\ifthenelse{\isundefined{\corollary}}{\newtheorem{corollary}{Corollary}}{}
\ifthenelse{\isundefined{\alg}}{\newtheorem{alg}{Algorithm}}{}
\ifthenelse{\isundefined{\example}}{\newtheorem{example}{Example}}{}

\newif\ifshowcomments
\ifshowcomments
\newcommand\pl[1]{\textcolor{red}{[PL: #1]}}
\newcommand\sidaw[1]{\textcolor{blue}{[sidaw: #1]}}
\newcommand\TODO[1]{\textcolor{red}{[TODO: #1]}}
\newcommand\cdm[1]{\marginpar{\footnotesize\raggedright\textcolor{blue}{#1}}}
\else
\newcommand\pl[1]{}
\newcommand\sidaw[1]{}
\newcommand\TODO[1]{}
\newcommand\cdm[1]{}
\fi

\newcommand{\autocap}[1]{
  \ifnum\spacefactor=3000
    \expandafter\MakeUppercase\fi%
  #1}

\newcommand\settingName{ILLG} 
\newcommand\gameName{SHRDLURN} 
\newcommand\metricName{\autocap{online accuracy}}

\newcommand\metricNames{\autocap{online accuracies}}

\newcommand\human{\autocap{human}} 
\newcommand\comp{\autocap{computer}} 

\newcommand\spammers{\autocap{spam players}}
\newcommand\spammer{\autocap{spam player}}

\newcommand{\wall}{\ensuremath{\sY}} 

\providecommand{\cond}{\ |\ }
\providecommand{\T}{\mathsf{T}} 

\providecommand{\utt}[1]{\textit{`#1'}}
\providecommand{\logf}[1]{\texttt{#1}}

\providecommand{\gram}{{\ensuremath{\rightarrow} }}
\usepackage{stmaryrd}

\providecommand{\means}{\ensuremath{\mapsto}}

\newcommand{\reg}[1]{||#1||_1}

\newcommand{\forcecommand}[3][0]{
  \providecommand{#2}{}
  \renewcommand{#2}[#1]{#3}
}
\forcecommand {\maximize} {\operatorname{maximize}}
\forcecommand {\minimize} {\operatorname{minimize}}

\forcecommand {\subjecto} {\operatorname{subject\ to}}
\forcecommand{\lone} {\ensuremath{\ell_1}}
\forcecommand{\ltwo} {\ensuremath{\ell_2}}

\newcommand\exec[2]{\ensuremath{\llbracket#2\rrbracket_{#1}}}
\newcommand\zremovered{z_{\logf{rm-red}}}
\newcommand\zremovecyan{z_{\logf{rm-cyan}}}

\aclfinalcopy 
\def\aclpaperid{696}
\title{Learning Language Games through Interaction}

\author{
  Sida I. Wang \qquad Percy Liang \qquad Christopher D. Manning\\
  Computer Science Department \\
  Stanford University \\
  {\tt \{sidaw,pliang,manning\}@cs.stanford.edu}
}
\date{}
\begin{document}
\maketitle

\begin{abstract}

We introduce a new language learning setting relevant to
building adaptive natural language interfaces. It is inspired by Wittgenstein's language games: a
human wishes to accomplish some task (e.g., achieving a certain
configuration of blocks), but can only communicate with a computer,
who performs the actual actions (e.g., removing all red blocks).
The computer initially knows nothing about language and therefore must
learn it from scratch through interaction, while the human adapts to the computer's capabilities.
We created a game called SHRDLURN in a blocks
world and collected interactions from 100 people playing it. 
First, we analyze
the humans' strategies, showing that using compositionality and avoiding
synonyms correlates positively with task performance.
Second, we compare computer strategies, showing
that modeling pragmatics on a semantic parsing model
accelerates learning for more strategic players.

\end{abstract}
\section{Introduction}
\begin{figure}[t]
\begin{center} 
\includegraphics[width=0.5\textwidth]{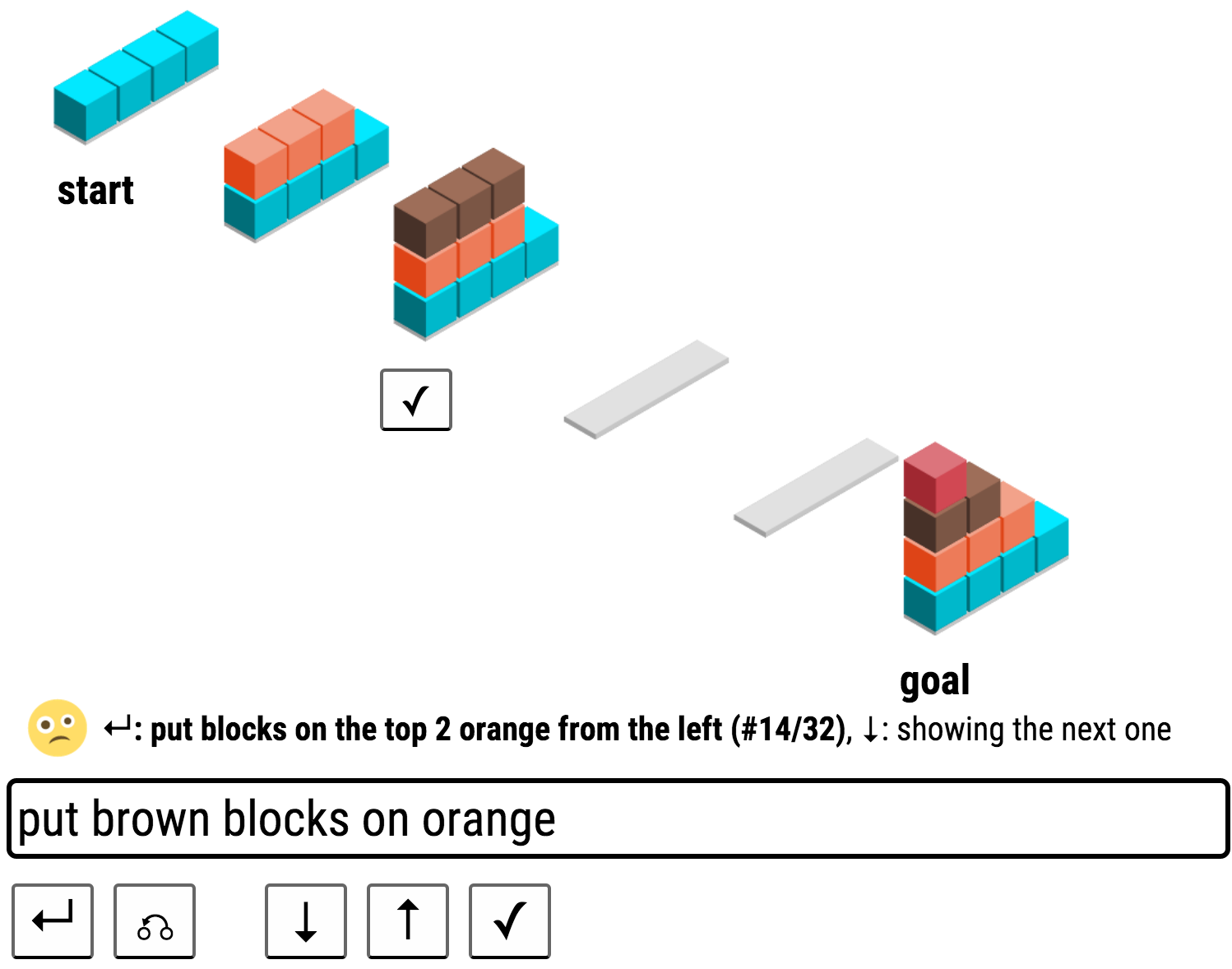}
\end{center}
\caption{\label{fig:setting}
  The \gameName{} game:
  the objective is to transform the start state into the goal state.
  The human types in an utterance, and the \comp{} (which does not know the goal state)
  tries to interpret the utterance
  and perform the corresponding action.
  The \comp{} initially knows nothing about the language,
  but through the human's feedback, learns the human's language
  while making progress towards the game goal.}
\end{figure}

\citet{wittgenstein1953philosophical} famously said that \emph{language derives
its meaning from use}, and introduced the concept of \emph{language games}
to illustrate the fluidity and purpose-orientedness of language.
He described how a builder B and an assistant A can use a primitive language
consisting of four words---\utt{block}, \utt{pillar}, \utt{slab},
\utt{beam}---to successfully communicate what block to pass from A to B.
This is only one such language; many others would also work for accomplishing
the cooperative goal.

This paper operationalizes and explores the idea of language games in a
learning setting,
which we call \emph{interactive learning through language games} (\settingName{})\@.
In the \settingName{} setting,
the two parties do not initially speak a common language,
but nonetheless need to collaboratively accomplish a goal.
Specifically, we created a game called \gameName{},\footnote{Demo: \url{http://shrdlurn.sidaw.xyz}} 
in homage to the 
seminal work of \citet{winograd1972language}.
As shown in \reffig{setting},
the objective is to transform a start state into a goal state,
but the only action the \human{} can take is entering an utterance.
The \comp{} parses the utterance and produces a ranked list of possible interpretations according to its
current model.  The \human{} scrolls through the list and chooses
the intended one, simultaneously advancing the state of the blocks
and providing feedback to the \comp{}.
Both the \human{} and the \comp{} wish to reach the goal state (only known to the \human{})
with as little scrolling as possible.
For the \comp{} to be successful,
it has to learn the \human{}'s language quickly over the course of the game,
so that the \human{} can accomplish the goal more efficiently.
Conversely, the \human{} must also accommodate the \comp{},
at least partially understanding what it can and cannot do.

We model the \comp{} in the \settingName{} as a semantic parser (\refsec{model}),
which maps natural language utterances (e.g., \utt{remove red}) into 
logical forms (e.g., $\logf{remove}(\logf{with}(\logf{red}))$).
The semantic parser has no seed lexicon and no annotated logical forms,
so it just generates many candidate logical forms.
Based on the \human{}'s feedback, it performs online gradient updates on the
parameters corresponding to simple lexical features.

During development,
it became evident that
while the \comp{} was eventually able to learn the language,
it was learning less quickly than one might hope.
For example, after learning that \utt{remove red} maps to \logf{remove(with(red))},
it would think that \utt{remove cyan} also mapped to \logf{remove(with(red))},
whereas a human would likely use mutual exclusivity to rule out that hypothesis
\citep{markman1988exclusivity}.
We therefore introduce a pragmatics model in which the \comp{} explicitly
reasons about the \human{},
in the spirit of previous work on pragmatics
\citep{golland10pragmatics,frank2012pragmatics,smith2013pragmatics}.
To make the model suitable for our \settingName{} setting,
we introduce a new online learning algorithm.
Empirically, we show that our pragmatic model improves
the \metricName{} by 8\% compared to our best non-pragmatic model on the 10 most successful
players (\refsec{results}).

What is special about the \settingName{} setting is the real-time nature of learning,
in which the human also learns and adapts to the \comp{}.
While the human can teach the \comp{} any language---English, Arabic, Polish, a
custom programming language---a good human player will choose to use utterances that
the \comp{} is more likely to learn quickly.
In the parlance of communication theory, the human \emph{accommodates} the \comp{}
\citep{giles2008communication,ireland2011language}.
Using Amazon Mechanical Turk, we collected and analyzed around 10k
utterances from 100 games of \gameName{}.
We show that successful players 
tend to use compositional utterances with a consistent vocabulary and syntax,
which matches the inductive biases of the \comp{}
(\refsec{analysis}). 
In addition, through this interaction, many players adapt to the computer by becoming more consistent,
more precise, and more concise.

On the practical side, natural language systems 
are often trained once and deployed, and users must
live with their imperfections.
We believe that studying the \settingName{} setting
will be integral for creating adaptive and customizable systems,
especially for resource-poor languages
and new domains where starting from close to scratch
is unavoidable.

\section{Setting}
\label{sec:setup}
We now describe the interactive learning of language games (\settingName{}) setting 
formally.\cdm{Is it bad that $\wall$ isn't defined until the next page?}
There are two players, the \human{} and the \comp{}.
The game proceeds through a fixed number of levels.
In each level, both players are presented with a starting state
$s \in \wall$, but only the \human{} sees the goal state $t \in \wall$.
(e.g. in \gameName{}, $\wall$ is the set of all configurations of blocks).
The \human{} transmits an utterance $x$
(e.g., \utt{remove red})
to the \comp{}.
The \comp{} then constructs a ranked list of candidate actions
$Z = [z_1, \dots, z_K] \subseteq \sZ$
(e.g., \logf{remove(with(red))}, \logf{add(with(orange))}, etc.),
where $\sZ$ is all possible actions.
For each $z_i \in Z$, it computes $y_i = \exec{s}{z_i}$, the successor
state from executing action $z_i$ on state $s$.
The \comp{} returns to the \human{}
the ordered list $Y = [y_1, \dots, y_K]$ of successor states.
The \human{} then chooses $y_i$ from the list $Y$
(we say the \comp{} is \emph{correct} if $i=1$).
The state then updates to $s = y_i$.
The level ends when $s = t$,
and the players advance to the next level.



Since only the \human{} knows the goal state $t$ and only the \comp{} can perform actions,
the only way for the two to play the game successfully is for the \human{} to
somehow encode the desired action in the utterance $x$.
However, we assume the two players do not have a shared language,
so the \human{} needs to pick a language and teach it to the \comp{}.
As an additional twist, the \human{} does not know the exact set of actions $\sZ$
(although they might have some preconception of the \comp{}'s
capabilities).%
\footnote{This is often the case when we try to interact with a new
software system or service before reading the manual.}
Finally, the \human{} only sees the outcomes of the \comp{}'s actions,
not the actual logical actions themselves.

We expect the game to proceed as follows:
In the beginning, the \comp{} does not understand what the \human{} is saying
and performs arbitrary actions.  As the \comp{} obtains feedback and learns,
the two should become more proficient at communicating and thus playing the game.
Herein lies our key design principle:
\emph{language learning should be necessary for the players to achieve good
game performance}.



\begin{table*}
  \begin{center}
\begin{tabular}{lll}
\textbf{Rule} & \textbf{Semantics} & \textbf{Description} \\[1ex]
Set                 & $\logf{all}()$ & all stacks \\
Color               & $\logf{cyan|brown|red|orange}$ & primitive color \\
Color \gram Set     & $\logf{with}(c)$ & stacks whose top block has color $c$ \\
Set \gram Set       & $\logf{not}(s)$ & all stacks except those in $s$ \\
Set \gram Set       & $\logf{leftmost|rightmost}(s)$ & leftmost/rightmost stack in $s$ \\
Set Color \gram Act & $\logf{add}(s,c)$ & add block with color $c$ on each stack in $s$ \\
Set \gram Act       & $\logf{remove}(s)$ & remove the topmost block of
                                           each stack in $s$ \\
\end{tabular}
\caption{The formal grammar defining the compositional action space
  $\sZ$ for \gameName{}. We use $c$ to denote a Color, and $s$ to
  denote a Set. 
For example, one action
that we have in \gameName{} is:
\utt{add an orange block to all 
but the leftmost brown block} \means{}
 \logf{add(not(leftmost(with(brown))),orange)}.
} 
\label{tab:grammar}
\end{center}
\end{table*}

\paragraph{\gameName{}.}
Let us now describe the details of our specific game, \gameName{}.
Each state $s \in \wall$ consists of stacks of colored blocks arranged
in a line (\reffig{setting}), where each stack is a vertical column of blocks.
The actions $\sZ$ are defined compositionally via the grammar in
\reftab{grammar}.  Each action either adds to or removes from a set of stacks,
and a set of stacks is computed via various set operations and selecting by color.
For example, the action $\logf{remove}(\logf{leftmost}(\logf{with}(\logf{red})))$
removes the top block from the leftmost stack whose topmost block is red.
The compositionality of the actions gives the \comp{} non-trivial capabilities.
Of course, the \human{} must teach a language to harness those capabilities,
while not quite knowing the exact extent of the capabilities. The
actual game proceeds according to a curriculum,
where the earlier levels only
need simpler actions with fewer predicates.


We designed \gameName{} in this way for several reasons.
First, visual block manipulations are intuitive and can
be easily crowdsourced, and it can be fun as an actual game that people would play.
Second, the action space is designed to be compositional,
mirroring the structure of natural language.
Third, many actions $z$ lead to the same successor state $y = \exec{s}{z}$;
e.g., the \utt{leftmost stack} might coincide with the \utt{stack with red blocks}
for some state $s$ and therefore an action involving either one would result in
the same outcome.
Since the \human{} only points out the correct $y$,
the \comp{} must grapple with this indirect supervision,
a reflection of real language learning.




\section{Semantic parsing model}
\label{sec:model}
Following \citet{zettlemoyer05ccg} and most recent work on semantic parsing,
we use a log-linear model over logical forms (actions) $z \in \sZ$
given an utterance $x$:
\begin{align}
p_\theta(z \cond x) &\propto \exp(\theta^\T \phi(x,z)),
\end{align}
where $\phi(x,z) \in \R^d$ is a feature vector and $\theta \in \R^d$ is a
parameter vector.
The denotation $y$ (successor state) is obtained by executing $z$ on a state $s$; formally, $y = \exec{s}{z}$.

\paragraph{Features.}
\label{sec:features}
Our features are $n$-grams (including skip-grams) conjoined
with tree-grams on the logical form side.
Specifically, on the utterance side (e.g., \utt{stack red on orange}),
we use unigrams ($\utt{stack}, *, *$),
bigrams ($\utt{red}, \utt{on}, *$),
trigrams ($\utt{red}, \utt{on}, \utt{orange}$),
and skip-trigrams ($\utt{stack}, *, \utt{on}$).
On the logical form side,
features corresponds to the predicates in the logical forms and their arguments.
For each predicate $h$, let $h.i$ be the $i$-th argument of $h$.
Then, we define \emph{tree-gram} features $\psi(h, d)$ for predicate $h$ and depth $d=0,1,2,3$
recursively as follows:
\begin{align*}
\psi(h,0) &= \{h\}, \\
\psi(h,d) &= \{(h, i, \psi(h.i,d-1)) \mid i=1,2,3\}.
\end{align*}
The set of all features is just the cross product of utterance
features and logical form features.
For example, if $x = \utt{enlever tout}$ and $z = \logf{remove(all())}$,
then features include:
\begin{quote}
\small
\begin{tabular}{ll}
$(\utt{enlever}, \logf{all})$                      & $(\utt{tout}, \logf{all})$                       \\
$(\utt{enlever}, \logf{remove})$                   & $(\utt{tout}, \logf{remove})$                    \\
$(\utt{enlever}, (\logf{remove}, 1, \logf{all}))$  &   \\
$(\utt{tout}, (\logf{remove}, 1, \logf{all}))$
\end{tabular}
\end{quote}

Note that we do not model an explicit alignment or derivation compositionally
connecting the utterance and the logical form,
in contrast to most traditional work in semantic parsing
\citep{zettlemoyer05ccg,wong07synchronous,liang11dcs,kwiatkowski10ccg,berant2013freebase},
instead following a looser model of semantics
similar to \citep{pasupat2015compositional}.
Modeling explicit alignments or derivations
is only computationally feasible when
we are learning from annotated logical forms or have a seed lexicon,
since the number of derivations is much larger than the number of logical forms.
In the \settingName{} setting, neither are available.

\paragraph{Generation/parsing.}

We generate logical forms from smallest to largest using beam search.
Specifically, for each size $n = 1, \dots, 8$,
we construct a set of logical forms of size $n$
(with exactly $n$ predicates)
by combining logical forms of smaller sizes according to the grammar rules in \reftab{grammar}.
For each $n$, we keep the $100$ logical forms $z$ with the highest score $\theta^\T \phi(x,z)$
according to the current model~$\theta$.
Let $Z$ be the set of logical forms on the final beam, which contains
logical forms of all sizes $n$.
During training,
due to pruning at intermediate sizes,
$Z$ is not guaranteed to contain the logical form that obtains the observed state $y$.
To mitigate this effect,
we use a curriculum so that only simple actions are needed in the initial
levels, giving the \human{} an opportunity to teach the \comp{} about basic
terms such as colors first before moving to larger composite actions.

The system executes all of the logical forms on the final beam $Z$,
and orders the resulting denotations $y$ by the maximum probability of any logical form
that produced it.\footnote{
We tried ordering based on the sum of the probabilities (which corresponds
to marginalizing out the logical form),
but this had the degenerate effect of assigning too much probability
mass to $y$ being the set of empty stacks,
which can result from many actions.}

\paragraph{Learning.}
When the \human{} provides feedback in the form of a particular $y$,
the system forms the following loss function:
\begin{align}
  \label{eqn:loss}
  \ell(\theta, x, y) &= -\log p_{\theta}(y \mid x, s) + \lambda \reg{\theta}, \\
  p_\theta(y \cond x, s) &= \sum_{z : \exec{s}{z} = y} p_\theta(z \cond x).
\end{align}
Then it makes a single gradient update using AdaGrad \citep{duchi10adagrad},
which maintains a per-feature step size.


\section{Modeling pragmatics}
\label{sec:pragmatics}
In our initial experience with the semantic parsing model described in \refsec{model},
we found that it was able to learn reasonably well,
but lacked a reasoning ability that one finds in human learners.
To illustrate the point,
consider the beginning of a game when $\theta=0$ in the log-linear model
$p_\theta(z \mid x)$.
Suppose that \human{} utters \utt{remove red} and then 
identifies $\zremovered = \logf{remove}(\logf{with}(\logf{red}))$ as the correct logical form.
The \comp{} then performs a gradient update on the loss function \refeqn{loss},
upweighting features such as
$(\utt{remove}, \logf{remove})$ and $(\utt{remove}, \logf{red})$.

Next, suppose the \human{} utters \utt{remove cyan}.
Note that $\zremovered$ will score higher than all other formulas
since the $(\utt{remove}, \logf{red})$ feature will fire again.
While statistically justified,
this behavior fails to meet our intuitive expectations for a smart language learner.
Moreover, this behavior is not specific to our model,
but applies to any statistical model that simply tries to fit the data
without additional prior knowledge about the specific language. 
While we would not expect the \comp{} to magically guess $\utt{remove
cyan} \means\logf{remove}(\logf{with}(\logf{cyan}))$,
it should at least push down the probability of $\zremovered$ because $\zremovered$
intuitively is already well-explained by another utterance \utt{remove red}.

This phenomenon, \emph{mutual exclusivity}, was studied by 
\citet{markman1988exclusivity}.
They found that children,
during their language acquisition process, reject a second label
for an object and treat it instead as a label for a novel object.




\paragraph{The pragmatic \comp{}.}
To model mutual exclusivity formally,
we turn to probabilistic models of pragmatics
\citep{golland10pragmatics,frank2012pragmatics,smith2013pragmatics,goodman2015prob},
which operationalize the ideas of \citet{grice75maxims}.
The central idea in these models is to
treat language as a cooperative game between a speaker (\human{}) and a listener (\comp{}) as we are doing,
but where the listener has an explicit model of the speaker's strategy,
which in turn models the listener.
Formally, let $S(x \cond z)$ be the speaker's strategy and $L(z \cond x)$ be the listener's strategy.
The speaker takes into account the literal semantic parsing model $p_\theta(z \mid x)$
as well as a prior over utterances $p(x)$,
while the listener considers the speaker $S(x \mid z)$ and a prior $p(z)$:
\begin{align}
\label{eqn:prag}
  S(x \cond z) &\propto \left(p_\theta(z \mid x) p(x)\right)^\beta, \\
  L(z \cond x) &\propto S(x \cond z) p(z),
\label{eqn:pragl}
\end{align}
where $\beta \ge 1$ is a hyperparameter that sharpens the distribution
\citep{smith2013pragmatics}.
The \comp{} would then use $L(z \cond x)$ to rank candidates rather than $p_\theta$.
Note that our pragmatic model only affects the ranking of actions
returned to the \human{} and does not affect the gradient updates of
the model $p_\theta$.


\begin{table}[t]
\begin{tabular}{ c | l l l  } 
& $\zremovered$ & $\zremovecyan$ & \multicolumn{1}{l}{$z_3,  z_4, \ldots$}\\ \hline
& \multicolumn{3}{c}{$p_{\theta}(z\mid x)$} \\
 \hline
 $\utt{remove red}$   & 0.8 & 0.1 & 0.1 \\ \hdashline
 $\utt{remove cyan}$ & \textcolor{red}{\bf 0.6} & 0.2 & 0.2 \\ \hline
& \multicolumn{3}{c}{$S(x\mid z)$} \\
 \hline
 $\utt{remove red}$   & 0.57 & 0.33 & 0.33 \\ \hdashline
 $\utt{remove cyan}$ & 0.43 & 0.67 & 0.67 \\ \hline
& \multicolumn{3}{c}{$L(z \mid x)$} \\
 \hline
 $\utt{remove red}$   & 0.46 & 0.27 & 0.27 \\ \hdashline
 $\utt{remove cyan}$ & 0.24 & \textcolor{blue}{\bf 0.38} & \textcolor{blue}{\bf 0.38} \\ 
\end{tabular}
\caption{ \label{tab:example1}
Suppose the \comp{} saw one example of \utt{remove red} \means $\zremovered$,
and then the \human{} utters \utt{remove cyan}.
{\bf top}: the literal listener, $p_\theta(z \cond x)$, mistakingly
chooses $\zremovered$ over $\zremovecyan$.
{\bf middle}: the pragmatic speaker, $S(x \cond z)$, assigns a higher
probability to to \utt{remove cyan} given $\zremovecyan$;
{\bf bottom}: the pragmatic listener, $L(z \cond x)$
 correctly assigns a lower probability to $\zremovered$ where $p(z)$
 is uniform.}
\end{table}

Let us walk through a simple example to see the effect of modeling pragmatics.
\reftab{example1} shows that the literal listener $p_\theta(z \cond x)$ assigns
high probability to $\zremovered$ for both \utt{remove red} and
\utt{remove cyan}.
Assuming a uniform $p(x)$ and $\beta=1$, the pragmatic speaker $S(x \cond z)$
corresponds to normalizing each column of $p_\theta$.
Note that if the pragmatic speaker wanted to convey $\zremovecyan$,
there is a decent chance that they would favor \utt{remove cyan}.
Next, assuming a uniform $p(z)$, the pragmatic listener $L(z \cond x)$
corresponds to normalizing each row of $S(x \cond z)$.
The result is that conditioned on \utt{remove cyan},
$\zremovecyan$ is now more likely than $\zremovered$,
which is the desired effect.

The pragmatic listener models the speaker as
a cooperative agent
who behaves in a way to maximize communicative success.
Certain speaker behaviors such as
avoiding synonyms (e.g., not \utt{delete cardinal}) and using a
consistent word ordering (e.g, not \utt{red remove})
fall out of the game theory.\footnote{
  Of course, synonyms and variable word order occur in real language.
  We would need a more complex game compared to \gameName{}
  to capture this effect.
}
For speakers that do not follow this strategy,
our pragmatic model is incorrect,
but as we get more data through game play,
the literal listener $p_\theta(z \cond x)$ will sharpen,
so that the literal listener and the pragmatic listener
will coincide in the limit.

\paragraph{Online learning with pragmatics.}
\SetAlCapSkip{0.6em}
\providecommand\algf{}
\begin{algorithm}
  \DontPrintSemicolon
  $\forall z, C(z) \gets 0$ \;
  $\forall z, Q(z) \gets \epsilon$ \;
  \Repeat{game ends}{
  {\algf receive utterance $x$ from human}
  $L(z \mid x) \propto \frac{P(z)}{Q(z)} p_{\theta}(z \mid x)^\beta $\; 
  {\algf send human a list $Y$ ranked by $L(z \mid x)$}\;\BlankLine
  {\algf receive $y \in Y$ from human\;}
  $\theta \gets \theta - \eta \nabla_{\theta}\ell(\theta, x, y)$\; 
  $Q(z) \gets Q(z) + p_{\theta}(z \mid x)^{\beta}$\;
  $C(z) \gets C(z) + p_{\theta}(z \mid x,\exec{s}{z}=y)$\;
  $P(z) \gets \frac{C(z) + \alpha}{\sum_{z' : C(z') > 0}
    \big(C(z')+\alpha\big)}$}
\caption{\label{alg:onlineprag} Online learning algorithm that updates the parameters of the semantic parser $\theta$
as well as counts $C,Q$ required to perform pragmatic reasoning.}
\end{algorithm}


To implement the pragmatic listener as defined in \refeqn{pragl},
we need to compute the speaker's normalization constant $\sum_x
p_{\theta}(z \mid x) p(x)$ in order to compute $S(x\mid z)$ in \refeqn{prag}.
This requires parsing all utterances $x$ based on
$p_\theta(z \cond x)$.
To avoid this heavy computation in an online setting,
we propose \refalg{onlineprag}, where some approximations
are used for the sake of efficiency.
First, to approximate the intractable sum over all utterances $x$,
we only use the examples that are seen to compute the normalization constant
$\sum_x p_{\theta}(z \mid x) p(x) \approx \sum_i p_{\theta}(z \mid
x_i)$.  Then, in order to avoid parsing all previous examples again using the current
parameters for each
new example, we store $Q(z) = \sum_i p_{\theta_i}(z
\mid x_i)^\beta$, where
$\theta_i$ is the parameter after the model updates on the $i^{th}$
example $x_i$.
While $\theta_i$ is different from the current parameter $\theta$,
$p_\theta(z \cond x_i) \approx p_{\theta_i}(z \cond x_i)$ for the
relevant example $x_i$, which is accounted for by both $\theta_i$ and $\theta$.

In \refalg{onlineprag}, the pragmatic listener $L(z \mid x)$ can be interpreted as an importance-weighted
version of the sharpened literal listener $p_{\theta}^\beta$, where it
is downweighted by $Q(z)$, which reflects which $z$'s
the literal listener prefers, and upweighted by $P(z)$, which is just
a smoothed estimate of the actual distribution over logical forms $p(z)$.
By construction, \refalg{onlineprag} is the same as \refeqn{prag} except that it uses
the normalization constant $Q$ based on stale parameters $\theta_i$
after seeing example, and it uses samples to compute the sum over $x$.
Following \refeqn{pragl}, we also need $p(z)$,
which is estimated by $P(z)$ using add-$\alpha$ smoothing on the
counts $C(z)$.
Note that $Q(z)$ and $C(z)$ are
updated \emph{after} the model parameters are updated for the current
example. 

Lastly, there is a small complication due to only observing the denotation $y$ and not the logical form $z$.
We simply give each consistent logical form
$\{z \mid \exec{s}{z}=y\}$ a
pseudocount based on the model: $C(z) \gets C(z) + p_{\theta}(z \mid
x,\exec{s}{z}=y)$ where $p_{\theta}(z \mid x,\exec{s}{z}=y) \propto
\exp(\theta^\T \phi(x,z))$ for $\exec{s}{z}=y$ (0 otherwise).

Compared to prior work where the setting is specifically
designed to require pragmatic inference,
pragmatics arises naturally in \settingName{}.
We think that this form of pragmatics is the most important during learning, and 
becomes less important if we had more data.
Indeed, if we have a lot of data and a small
number of possible $z$s, then
$L(z|x)  \approx p_{\theta}(z|x)$ as $\sum_x p_{\theta}(z|x) p(x)
\rightarrow p(z)$ when $\beta=1$.%
\footnote{Technically, we also need $p_{\theta}$ to be \emph{well-specified}. }
However, for semantic parsing, we would not be in this regime even if
we have a large amount of training data.
In particular, we are nowhere near that regime in \gameName{},
and most of our utterances / logical forms are seen only once, and the importance of modeling pragmatics remains.

\section{Experiments}
\label{sec:experiments}
\subsection{Setting}
\label{sec:settings}
\paragraph{Data.}
Using Amazon Mechanical Turk (AMT), we paid 100 workers 3 dollars each to play
\gameName{}.
In total, we have 10223 utterances along with their starting states
$s$. Of these, 8874 utterances are labeled with their denotations
$y$; the rest are unlabeled, since the player can try any
utterance without accepting an action.
100 players completed the entire game under identical
settings.
We deliberately chose to start from scratch for every worker,
so that we can study the diversity of strategies that different people
used in a controlled setting.

Each game consists of 50 blocks tasks divided into 5 levels of
10 tasks each, in increasing complexity.
Each level aims to reach an end goal given a start state.
Each game took on average 89 utterances
to complete.\footnote{
  This number is not 50 because some block tasks need multiple
  steps and players are also allowed to explore without reaching the goal.} 
It only took 6 hours to complete these 100 games on AMT and each game
took around an hour on average according to AMT's \emph{work time} tracker (which
does not account for multi-tasking players).
The players were provided minimal instructions on the game controls.
Importantly, we gave no example utterances in order to avoid biasing their language use.
Around 20 players were confused and told us that the instructions were
not clear and gave us mostly spam utterances.
Fortunately, most players understood the setting and some even
enjoyed \gameName{} as reflected by their optional comments:
{\it
\begin{itemize}[noitemsep]
\item That was probably the most fun thing I have ever done on mTurk.
\item Wow this was one mind bending games [sic]. 
\end{itemize}
}
\paragraph{Metrics.} 
We use the \emph{number of scrolls} as a measure of game
performance for each player. For each
example, the number of scrolls is the position in the list $Y$ of the action selected by the player.
It was possible to complete this version of \gameName{} by
scrolling (all actions can be found in the first 125 of $Y$)---22 of the 100
players failed to teach an actual language, and instead finished the game
mostly by scrolling. Let us call them
\emph{\spammer s}, who usually typed single
letters, random words, digits, or random phrases (e.g. \utt{how are you}).
Overall, \spammer s had to scroll a lot: 21.6 scrolls per utterance versus only
7.4 for the non-\spammer s. 

\subsection{Human strategies}
\label{sec:analysis}
\begin{table*}
\footnotesize 
  \begin{tabular}{p{0.3\textwidth} p{0.3\textwidth} p{0.3\textwidth}}
\multicolumn{3}{c}{\bf Most successful players (1st--20th)} \\ 
\midrule
\multirow{3}{0.3\textwidth} 
{rem cy pos 1,
stack or blk pos 4,
rem blk pos 2 thru 5,
rem blk pos 2 thru 4,
stack bn blk pos 1 thru 2,
fill bn blk,
stack or blk pos 2 thru 6,
rem cy blk pos 2 
fill rd blk  \textbf{(3.01)}} &
\multirow{3}{0.3\textwidth} 
{remove the brown block,
remove all orange blocks,
put brown block on orange blocks,
put orange blocks on all blocks,
put blue block on leftmost blue block in top row \textbf{ (2.78)}}&
\multirow{3}{0.3\textwidth} 
{Remove the center block, Remove the red block, Remove all red blocks,
Remove the first orange block, 
Put a brown block on the first brown block,
Add blue block on first blue block \textbf{(2.72)}}
\\ \\ & & \\ & & \\ & &\\\noalign{\vskip 2mm}  
\multicolumn{3}{c}{\bf Average players (21th--50th)} \\
\midrule 
\multirow{1}{0.3\textwidth} 
{reinsert pink,
take brown, put in pink,
remove two pink from second layer,
Add two red to second layer in odd intervals,
Add five pink to second layer,
Remove one blue and one brown from bottom layer  \textbf{(9.17)} }&
\multirow{1}{0.3\textwidth} 
{remove red, 
remove 1 red, 
remove 2 4 orange, 
add 2 red,
add 1 2 3 4 blue, 
emove 1 3 5 orange,
add 2 4 orange, 
add 2 orange, 
remove 2 3 brown, 
add 1 2 3 4 5 red,
remove 2 3 4 5 6, 
remove 2, 
add 1 2 3 4 6 red \textbf{(8.37)}
}&
\multirow{1}{0.3\textwidth}{
move second cube,
double red with blue,
double first red with red,
triple second and fourth with orange,
add red,
remove orange on row two,
add blue to column two,
add brown on first and third \textbf{(7.18)}
}
\\ \\ & & \\ & & \\ & &\\ && \\ \noalign{\vskip 2mm}  
\multicolumn{3}{c}{\bf Least successful players (51th--)} \\ 
\midrule 
\multirow{1}{0.3\textwidth} 
{holdleftmost,
holdbrown,
holdleftmost,
blueonblue,
brownonblue1,
blueonorange,
holdblue,
holdorange2,
blueonred2 ,
holdends1,
holdrightend,
hold2,
orangeonorangerightmost \textbf{(14.15)}} &
\multirow{1}{0.3\textwidth}{
`add red cubes on center left, center right, far left and far right',
`remove blue blocks on row two column two, row two column four',
remove red blocks in center left and center right on second row
\textbf{(12.6)}
}
&
\multirow{1}{0.3\textwidth}{
laugh with me,
red blocks with one aqua,
aqua red alternate,
brown red red orange aqua orange,
red brown red brown red brown,
space red orange red,
second level red space red space red space 
\textbf{(14.32)}
}
\\ \\ & & \\ & & \\ & &\\ \\ \noalign{\vskip 2mm}
\multicolumn{3}{c}{\bf Spam players ($\sim$ 85th--100)} \\ 
 \midrule 
\multicolumn{3}{c}{next, hello happy, how are you, move, gold, build goal blocks, 23,house, gabboli, x, run,,xav,
    d, j, xcv, dulicate goal (21.7)} 
\\  \noalign{\vskip 2mm}
\multicolumn{3}{c}{\bf Most interesting} \\ 
 \midrule
\multirow{4}{0.3\textwidth}
{\foreignlanguage{polish}{usuń brązowe klocki,
postaw pomarańczowy klocek na pierwszym klocku,
postaw czerwone klocki na pomarańczowych,
usuń pomarańczowe klocki w górnym rzędzie}} &
\multirow{4}{0.3\textwidth}
{rm scat 
+ 1 c,
+ 1 c, 
rm sh, 
+ 1 2 4 sh, 
+ 1 c, 
- 4 o,
rm 1 r, 
+ 1 3 o, 
full fill c, 
rm o, 
full fill sh, 
- 1 3, 
full fill sh, 
rm sh,
rm r, 
+ 2 3 r, 
rm o, 
+ 3 sh, 
+ 2 3 sh, 
rm b, 
- 1 o, 
+ 2 c, 
} &
\multirow{4}{0.3\textwidth}
{mBROWN,mBLUE,mORANGE
RED+ORANGE\^{}ORANGE,
BROWN+BROWNm1+BROWNm3,
ORANGE +BROWN +ORANGE\^{}m1+
ORANGE\^{}m3 + BROWN\^{}\^{}2 + BROWN\^{}\^{}4 }\\
\noalign{\vskip 18mm}
\end{tabular}
 \caption{
Example utterances, along with the average number of scrolls for that
player in parentheses. Success is measured by the number of scrolls,
where the more successful players need less scrolls.
1) The 20 most successful players tend to use consistent and concise language whose
   semantics is similar to our logical language.
2) Average players tend to be slightly
more verbose and inconsistent (left and right), or significantly different from our
logical langauge (middle). 
3) Reasons for being unsuccessful vary. Left: no tokenization, 
middle: used a coordinate system and many conjunctions;
right: confused in the beginning, and used a language very different from our logical language.}
\label{tab:examples}
\pl{put number of scrolls in bold or italics or something}
\pl{say successful means no. scrolls}
\end{table*}
Some example utterances can
be found in \reftab{examples}. Most of the players used English,
but vary in their adherence to conventions such as use of
determiners, plurals, and proper word ordering.
5 players invented their own language, which are more
precise, more consistent than general English. One player used Polish, and
another used Polish notation (bottom of \reftab{examples}).

\providecommand\changesto{\textbf{became}}
Overall, we find that many players adapt in \settingName{} by becoming
more consistent, less verbose, and more precise, even if they used
standard English at the beginning.
For example, some players became more consistent over time 
(e.g. from using both \utt{remove} and \utt{discard} to only using
\utt{remove}).
In terms of verbosity, removing
function words like determiners as the game progresses is a common
adaptation. In each of the following examples from different players, we compare an utterance
that appeared early in the game
to a similar utterance that appeared later: 
\utt{Remove the red ones} \changesto{} \utt{Remove red.}; 
\utt{add brown on top of red} \changesto{} \utt{add orange on red}; 
\utt{add red blocks to all red blocks} \changesto{}  \utt{add red to
  red}; 
\utt{dark red} \changesto{} \utt{red};
one player used \utt{the} in all of the first 20 utterances, and then never used \utt{the}
in the last 75 utterances.  

Players also vary in precision, ranging from overspecified (e.g. \utt{remove the
orange cube at the left},  \utt{remove red blocks from top row}) to underspecified or requiring context
(e.g. \utt{change colors}, \utt{add one blue}, 
\utt{Build more blocus}, \utt{Move the blocks fool},
\utt{Add two red cubes}). We found that some players became more precise
over time, as they gain a better understanding of \settingName{}.

Most players use utterances that actually do not match our logical
language in \reftab{grammar}, even the successful players.
In particular, numbers are often used. While
some concepts always have the same effect in our blocks world (e.g. \utt{first block} means \logf{leftmost}),
most are different.
More concretely,
of the top 10 players, 7 used numbers of some form and only 3 players matched our logical language.
Some players who did not match the logical language performed quite well nevertheless. One
possible explanation is because the action required is somewhat constrained
by the logical language and some tokens can have unintended interpretations.
For example, the \comp{} can
correctly interpret numerical positional references, as long as the player only
refers to the leftmost and rightmost positions.
So if the player says \utt{rem blk pos 4} and \utt{rem blk pos 1}, 
the \comp{} can interpret \utt{pos} as
\logf{rightmost} and interpret the bigram
$(\utt{pos},\utt{1})$ as \logf{leftmost}. 
On the other hand, players who deviated significantly by
describing the desired state declaratively (e.g. \utt{red orange red}, \utt{246}) rather than using actions,
or a coordinate system (e.g. \utt{row two column two}) performed poorly.
Although players do not have to match our
logical language exactly to perform well, being similar is definitely helpful.

\paragraph{Compositionality.} As far as we can tell,
all players used a compositional language;
no one invented unrelated words for each action.
Interestingly, 3 players did not put spaces between words.
Since we assume monomorphemic words separated by spaces,
they had to do a lot of scrolling as a result (e.g., 14.15 with
utterances like \utt{orangeonorangerightmost}).




\subsection{Computer strategies}
\label{sec:results}
We now present quantitative results on how
quickly the \comp{} can learn, where our goal is to achieve high accuracy on new
utterances as we make just a single pass over the data. The number of
scrolls used to evaluate player is sensitive to outliers and not as
intuitive as accuracy.
Instead, we consider \emph{\metricName{}}, described as follows.
Formally, if a player produced $T$ utterances $x^{(j)}$ and
labeled them $y^{(j)}$, then
\begin{align*}
\text{\metricName{}} \eqdef \frac1{T} \sum_{j=1}^T \1\left[y^{(j)} =
  \exec{s^{(j)}}{z^{(j)}}\right],
\end{align*}
where $z^{(j)} = \arg\max_z p_{\theta^{(j-1)}}(z | x^{(j)})$ is the model
prediction based  on the previous parameter $\theta^{(j-1)}$.
Note that the \metricName{} is defined with respect to the player-reported labels,
which only corresponds to the actual accuracy if the player is precise and honest.
This is not true for most \spammers{}. 

\label{sec:quant}
\begin{figure}[ht!!]
\centering

\begin{subfigure}{.23\textwidth}
  \centering
 \includegraphics[width=1\textwidth]{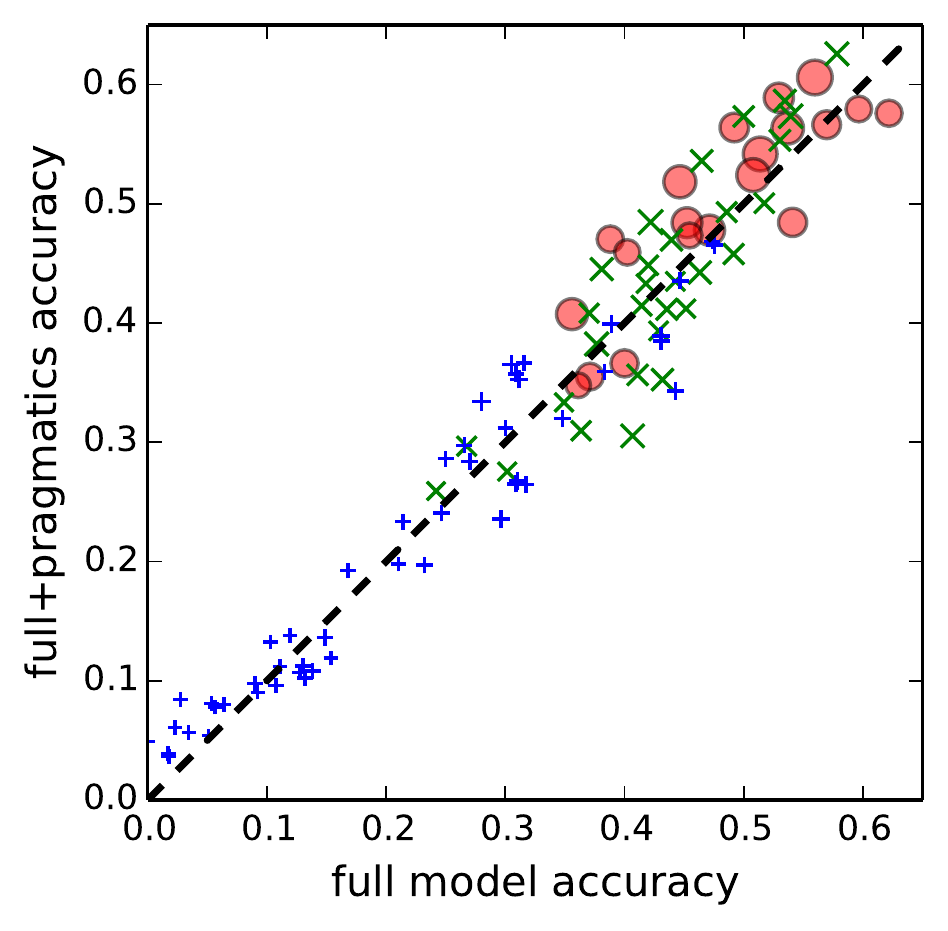}
  \caption{}
  \label{fig:pragvfull}
\end{subfigure}
\begin{subfigure}{.23\textwidth}
  \centering
 \includegraphics[width=1\textwidth]{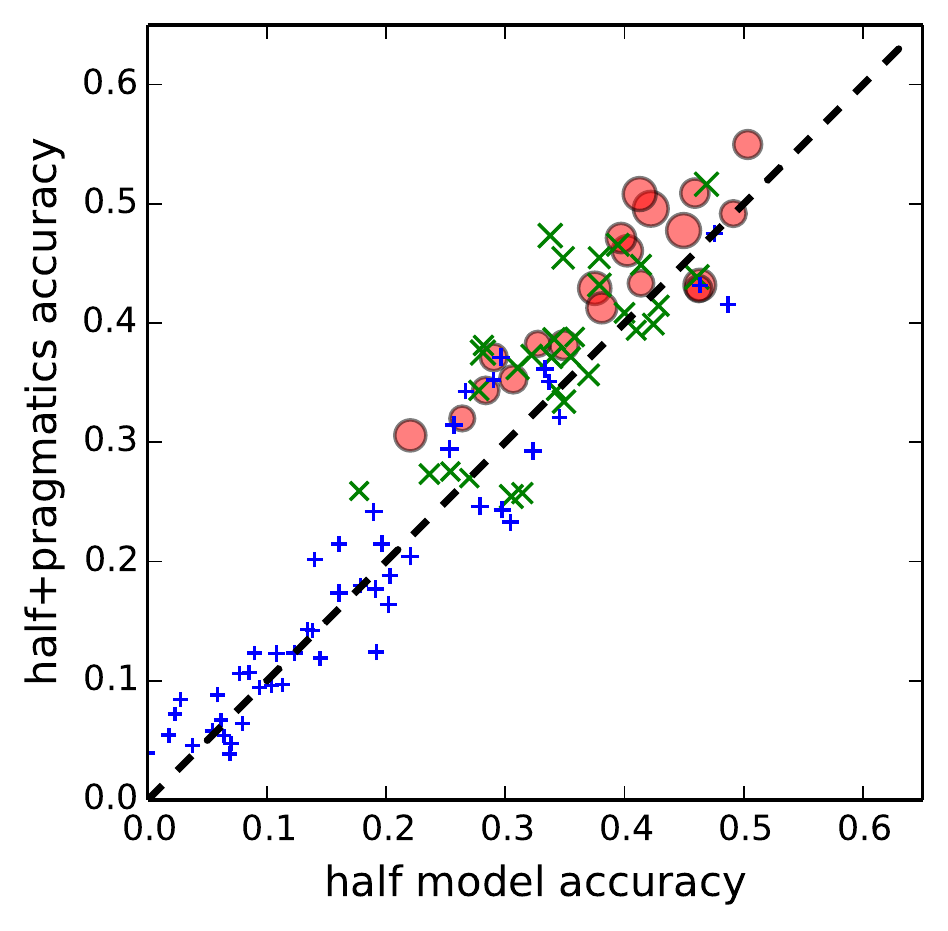}
  \caption{}
  \label{fig:pragvutt}
\end{subfigure}


\caption{
Pragmatics improve \metricName{}. In these plots, each marker is a player.
red o: players who ranked 1--20 in terms of minimizing number of scrolls, 
green x: players 20--50;
blue +: lower than 50 (includes \spammers{}).
Marker sizes correspond to player rank, where better players
are depicted with larger markers.
\textbf{\ref{fig:pragvfull}}: \metricNames{} with and without pragmatics on the full model;
\textbf{\ref{fig:pragvutt}}: same for the half model.
}
\label{fig:results}
\vspace{-0.5em}
\end{figure}

\begin{table}[ht]
\begin{tabular}{ c | l l l l } 
& \multicolumn{4}{c}{players ranked by \# of scrolls}\\ \hline
Method & top 10 & top 20 & top 50 & all 100 \\ \hline
memorize & 25.4 & 24.5 & 22.5 & 17.6 \\ \hdashline
half model& 38.7 & 38.4 & 36.0 & 27.0 \\
half + prag & 43.7& 42.7 & 39.7 & 29.4\\ \hdashline
full model & 48.6 & 47.8 & 44.9 & 33.3 \\
full + prag & 52.8 & 49.8 & 45.8 & 33.8\\
\end{tabular}
\caption{ \label{tab:results} Average \metricName{} under various settings.
memorize: featurize entire utterance and logical form non-compositionally;
half model: featurize the utterances with unigrams, bigrams, and skip-grams but conjoin with the entire logical form;
full model: the model described in \refsec{model};
+prag: the models above, with our online pragmatics algorithm described in \refsec{pragmatics}.
Both compositionality and pragmatics improve accuracy.
}
\end{table}

\paragraph{Compositionality.}
To study the importance of compositionality,
we consider two baselines.
First, consider a non-compositional model (\emph{memorize}) that just remembers pairs of complete utterance and
logical forms.
We implement this using indicator features on features
$(x,z)$, e.g., $(\utt{remove all the red blocks}, \zremovered)$,
and use a large learning rate.
Second, we consider a model (\emph{half}) that 
treats
utterances compositionally with unigrams, bigrams, and
skip-trigrams features,
but the logical forms are regarded as non-compositional,
so we have features such as
$(\utt{remove}, \zremovered)$, $(\utt{red}, \zremovered)$, etc.


\reftab{results} shows that the full model (\refsec{features})
significantly outperforms both the \emph{memorize} and \emph{half} baselines.
The learning rate  $\eta=0.1$ is selected via cross validation, and we
used $\alpha=1$ and $\beta=3$ following \citet{smith2013pragmatics}.

\paragraph{Pragmatics.}
Next, we study the effect of pragmatics on \metricName{}.
\reffig{results}
shows that modeling pragmatics helps successful players (e.g., top 10 by number of scrolls)
who use precise and consistent languages.
Interestingly, our pragmatics model did not help and can even hurt the less
successful players who are less precise and consistent.
This is expected behavior:
the pragmatics model assumes that the human is cooperative and behaving rationally.
For the bottom half of the players, this assumption is not true,
in which case the pragmatics model is not useful. 




\section{Related Work and Discussion}
Our work connects with a broad body of work on grounded language,
in which language is used in some environment as a means towards some goal.
Examples include playing games \citep{branavan09reinforcement,branavan10high,reckman2010virtualgame}
interacting with robotics \citep{tellex2011understanding,tellex2014asking},
and following instructions \citep{vogel10navigate,chen11navigate,artzi2013weakly}
Semantic parsing utterances to logical forms, which we leverage, plays an important role
in these settings
\citep{kollar2010grounding,matuszek2012grounded,artzi2013weakly}.

What makes this work unique is
our new interactive learning of language games (\settingName{}) setting,
in which a model has to learn a language from \emph{scratch} through interaction.
While online gradient descent is frequently used,
for example in semantic parsing \citep{zettlemoyer07relaxed,chen12lexicon},
we using it in a truly online setting, taking one pass over the data and
measuring online accuracy \citep{cesabianchi06prediction}.

To speed up learning,
we leverage computational models of pragmatics 
\citep{jaeger08game,golland10pragmatics,frank2012pragmatics,smith2013pragmatics,vogel2013emergence}.
The main difference is these previous works use pragmatics with a trained base model,
whereas we learn the model online.
\citet{monroe2015pragmatics} uses learning to improve the pragmatics model.
In contrast, we use pragmatics to speed up the learning process
by capturing phenomena like mutual exclusivity \citep{markman1988exclusivity}. 
We also differ from prior work in several details.
First, we model pragmatics in the online learning setting where we use
an online update for the pragmatics model. Second, unlikely the
reference games where pragmatic effects plays an important role by
design, \gameName{} is not specifically designed to require pragmatics. 
The improvement we get is mainly due to players trying to be
consistent in their language use.
Finaly, we treat both the utterance and the logical
forms as featurized compositional objects.
\citet{smith2013pragmatics} treats utterances (i.e. words) and logical
forms (i.e. objects) as categories;
\citet{monroe2015pragmatics} used features, but also over flat categories.
\pl{we should cite \citep{goodman2015prob} - they have pragmatics with compositional semantics}

Looking forward, we believe that the \settingName{} setting is worth
studying and has important implications for natural language interfaces.
Today, these systems are trained once and deployed.
If these systems could quickly adapt to user feedback in real-time as in this work,
then we might be able to more readily create
systems for resource-poor languages
and new domains, that are customizable and improve through use.

\section*{Acknowledgments}
DARPA Communicating with Computers (CwC) program under ARO
prime contract no. W911NF-15-1-0462. The first author is supported by
a NSERC PGS-D fellowship. In addition, we thank Will Monroe, and Chris Potts
for their insightful comments and discussions on pragmatics.
\section*{Reproducibility} All code, data, and experiments for this
paper are available on the CodaLab platform: \\
{\small \url{https://worksheets.codalab.org/worksheets/0x9fe4d080bac944e9a6bd58478cb05e5e}}\\
The client side code is here: \\
{\small \url{https://github.com/sidaw/shrdlurn/tree/acl16-demo}}\\
and a demo: {\small \url{http://shrdlurn.sidaw.xyz}}

\bibliography{refdb/all}


\end{document}